\documentclass{amia}
\usepackage{tabularx}
\usepackage{multirow}
\usepackage{booktabs}
\usepackage{graphicx}
\usepackage{makecell}
\usepackage{threeparttable}
\usepackage{caption}
\usepackage{xltabular}
\newcounter{supptable}

\usepackage[colorlinks=true, urlcolor=blue, linkcolor=red]{hyperref}
\begin{document}

\title{Information Extraction from Clinical Notes: Are We Ready to Switch to Large Language Models?}

\author{Yan Hu$^1$, Xu Zuo$^1$, Yujia Zhou$^2$, Xueqing Peng$^2$, Jimin Huang$^2$, Vipina K. Keloth$^2$, Vincent J. Zhang$^2$, Ruey-Ling Weng$^2$, Qingyu Chen$^2$, Xiaoqian Jiang$^1$, Kirk E. Roberts$^1$, Hua Xu$^2$ }

\institutes{
    $^1$ McWilliams School of Biomedical Informatics, The University of Texas Health Science Center at Houston, Houston, Texas, USA \\
    $^2$ Department of Biomedical Informatics and Data Science, Yale School of Medicine, Yale University, New Haven, USA \\
}

\maketitle

\section*{Abstract}

\textbf{Backgrounds:}\\ 
Information extraction (IE) is critical in clinical natural language processing (NLP). While large language models (LLMs) excel on generative tasks, their performance on extractive tasks remains debated.\\
\textbf{Methods:} \\
We investigated two fundamental clinical IE tasks: Named Entity Recognition (NER) and Relation Extraction (RE). Using 1,588 clinical notes from four data sources including UT Physicians (UTP), MTSamples, MIMIC-III, and i2b2, we developed a comprehensive annotated corpus covering 4 main clinical entities (problems, tests, medications, and other treatments) and 16 modifiers (e.g., negation, certainty). We instruction-tuned LLaMA-2 and LLaMA-3 for both tasks, comparing their performance, generalizability, computational resources, and throughput to BERT. \\
\textbf{Results:} \\
LLaMA models consistently outperformed BERT across datasets and tasks. With sufficient training data (e.g., UTP), LLaMA showed marginal improvements (1\% on NER, 1.5-3.7\% on RE) over BERT; for limited training data (e.g., MTSamples, MIMIC-III), the improvements were greater. On the unseen i2b2 dataset, LLaMA-3-70B outperformed BERT by over 7\% (F1) on NER and 4\% (F1) on RE. However, LLaMA models required significantly more computing resources (i.e., memory and GPUs) and ran up to 28 times slower than BERT. To facilitate LLM use in clinical IE, we implemented "Kiwi," a clinical IE package featuring both BERT and LLaMA models and made it available at \url{https://kiwi.clinicalnlp.org/.} \\
\textbf{Conclusion:} \\
This study is among the first to develop and evaluate a comprehensive clinical IE system using open-source LLMs. Results indicate that LLaMA models outperform BERT for clinical NER and RE but with higher computational costs and lower throughputs. These findings highlight that choosing between LLMs and traditional deep learning methods for clinical IE applications should remain task-specific, taking into account both performance metrics and practical considerations such as available computing resources and the intended use case scenarios.

\section*{1 Introduction}

The digital transformation of healthcare records into Electronic Health Records (EHRs) represents one of the most significant advancements in the modern healthcare system \cite{gopal_digital_2019}. This transformation has paved the way for approaches to advance patient care and clinical research \cite{bhavnani_2017_2017}. Among the data contained within EHRs, clinical notes have the most comprehensive and information-rich patient data \cite{dash_using_2020}. These narrative texts, written by healthcare professionals, capture a detailed picture of patient information, including diagnosis, treatments, patient outcomes, and much more \cite{zhang_five_2013}. Despite their potential, the massive, complex and unstructured nature of clinical notes has largely presented significant challenges for their effective use \cite{sheikhalishahi_natural_2019}.

Clinical information extraction (IE) aims to bridge the gap between unstructured text and structured data analysis \cite{kreimeyer_natural_2017}. By automating this process using Natural Language Processing (NLP) techniques such as Named Entity Recognition (NER) and Relation Extraction (RE), clinical IE offers solutions to these challenges \cite{bose_survey_2021,nasar_named_2022}. Over the past 20 years, most clinical NLP research has focused on clinical IE. For instance, 8 out of 10 challenges in the i2b2/n2c2 series from 2006 to 2019 have centered on clinical IE tasks \cite{uzuner_2010_2011,uzuner_evaluating_2012,uzuner_extracting_2010,uzuner_identifying_2008,uzuner_recognizing_2009,sun_evaluating_2013,henry_2018_2020,wang_2019_2020}. BERT-based models have previously demonstrated state-of-the-art (SOTA) performance in clinical IE tasks \cite{si_enhancing_2019,alsentzer_publicly_2019}. However, the complexity and variability of natural language used by different healthcare professionals present substantial barriers to effective clinical IE \cite{sheikhalishahi_natural_2019}. Systems developed using documents from a specific institution often do not generalize well when applied to clinical notes from other institutions or medical specialties \cite{wang_clinical_2018}.

Recently, large language models (LLMs) such as Generative Pre-trained Transformers (GPT) and the Large Language Model Meta AI (LLaMA) series \cite{yang_harnessing_2024,hadi_large_2024,myers_foundation_2024} have shown phenomenal capabilities in generative NLP tasks, including text summarization, question answering, and machine translation \cite{zhang_review_2022,chen_large_2023}. However, the application of these models to traditional IE tasks has yielded mixed results, with outcomes varying across different studies and scenarios, sparking debates within the research community. For example, in zero-/few-shot settings, LLMs often do not outperform traditional fine-tuned BERT models \cite{hu_improving_2024,liu_summary_2023,liu_large_2024}. While some research indicates that LLMs can achieve comparable or slightly better performance than traditional models after fine-tuning, these findings are frequently based on datasets from similar text types, raising questions about the models' generalizability across diverse text types \cite{dagdelen_structured_2024,goel_llms_2023}. Additionally, many studies focus on datasets with limited entity types or relation types \cite{keloth_advancing_2024,andrew_evaluating_2024,fornasiere_medical_2024}. To the best of our knowledge, no comprehensive study has compared fine-tuned LLMs and BERT models in terms of performance and generalizability across clinical notes from multiple data sources, entity types, and relation types. Moreover, there is a lack of research examining the trade-offs between model performance, throughput, and energy consumption, which is particularly crucial given the varying computational resources available in different hospitals and research institutions.

To address these challenges, we conducted an in-depth study to systematically compare LLM-based and BERT-based models on comprehensive clinical NER and RE tasks using a variety of evaluation metrics. Our study makes several key contributions to the clinical IE field. First, we developed one of the largest and most comprehensive cross-institutional annotated corpora for clinical NER and RE, comprising 1,588 clinical notes from multiple institutions. This corpus encompasses four primary clinical entities and 16 corresponding modifiers, providing a robust foundation for advancing clinical information extraction. Using the developed corpus, we systematically evaluated the performance of instruction-tuned LLaMA-3 models, demonstrating their superiority over previous state-of-the-art BERT-based models in clinical NER and RE tasks, particularly in low-resource and cross-institutional settings, with F1 score gains of up to 7\% for NER and 4\% for RE. Beyond performance metrics, we compared LLM-based and BERT-based clinical IE models across additional dimensions, including speed/throughput, computational resource requirements, and energy efficiency, which reveals some constraints of LLM-based IE models on real-world applications. Based on these findings, we provided detailed recommendations for model selection across diverse clinical NER and RE scenarios, bridging the gap between AI research and practical clinical use. Furthermore, we introduce Kiwi, one of the first publicly available LLM-based clinical IE systems. Kiwi provides both BERT-based and LLaMA-based models, allowing users to select the most appropriate tools based on their specific needs, computational resources, and task requirements, and it is freely accessible at \url{https://kiwi.clinicalnlp.org/}.

\section*{2 Methods}
\subsection*{2.1 Task}
Our study includes two clinical IE tasks: Named Entity Recognition (NER) and Relation Extraction (RE). NER identifies entity types and boundaries in text \cite{li_survey_2020}, while RE identifies relationships between entity pairs \cite{zhou_exploring_2005}. We instruction-tuned LLaMA-2 and LLaMA-3 models and compared them to fine-tuned BERT. We focused on extracting four main entities—Medical Problems, Tests, Drugs and other Treatments—along with 16 modifiers (e.g., Severity, Temporal, Dosage). In total, there are 21 possible relations between main entities and modifiers, as detailed in Table 1. Annotation methods and guidelines are available in the Supplementary File and GitHub repository.

\begin{figure}[H]
    \centering
    \fbox{\includegraphics[width=1.0\textwidth]{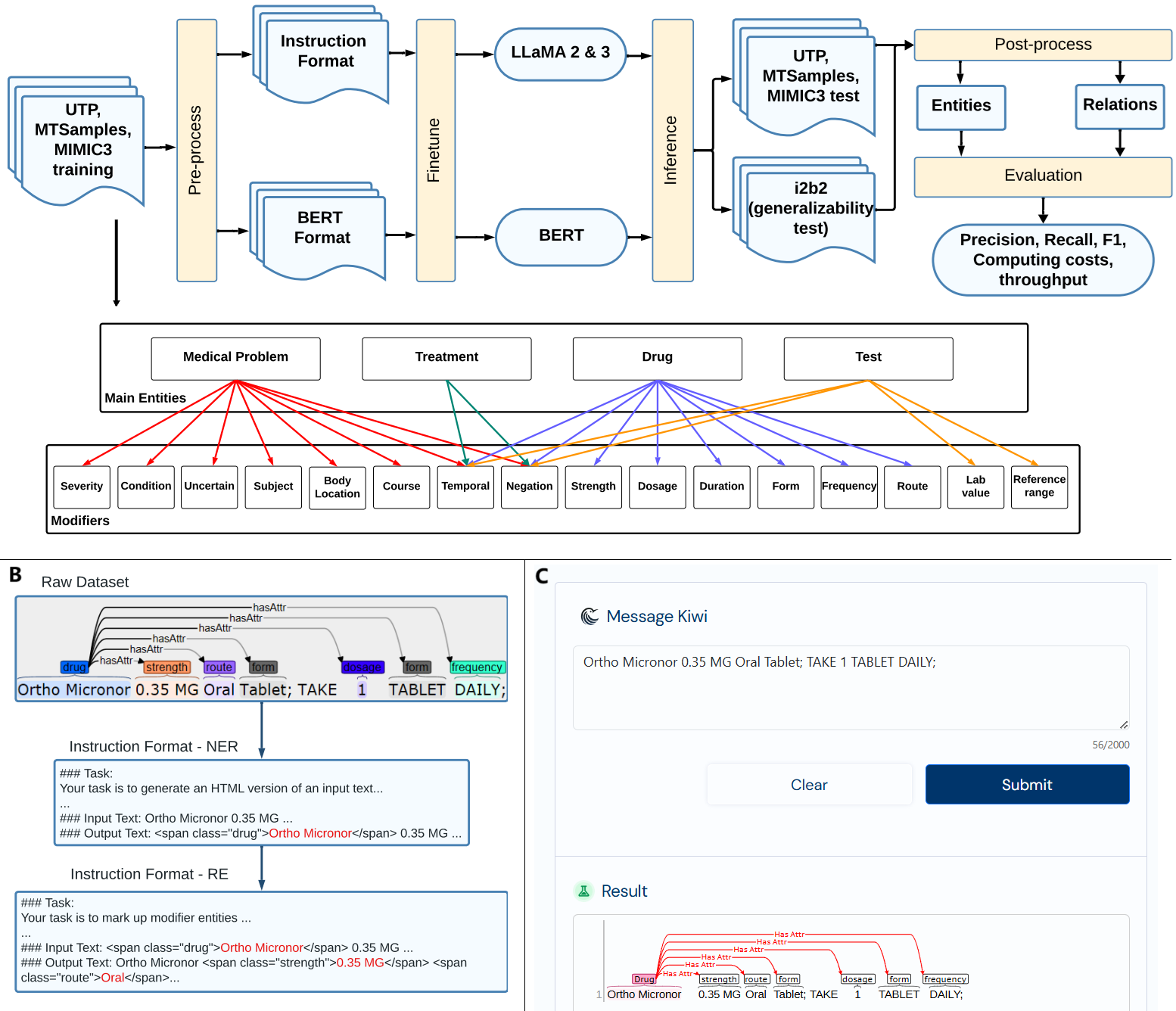}}
    \caption*{\footnotesize UTP: UT Physicians; MTSamples: Transcribed Medical Transcription Sample Reports and Examples; MIMIC-III: Medical Information Mart for Intensive Care; i2b2: Informatics for Integrating Biology \& the Bedside}
    \caption{(A) Overview of the study workflow, including entity and relation types in the dataset. (B) Examples of instructions for NER and RE, using a unified format with span tags to highlight all entities and modifiers. (C) The GUI of Kiwi online demo.}
\end{figure}

\subsection*{2.2 Dataset}
We developed a comprehensive clinical IE corpus that is not limited to specific diseases or treatments. It consists of real-world and synthetic clinical notes collected cross-institutionally, including 1,342 clinical notes from UT Physicians (UTP), 146 from Transcribed Medical Transcription Sample Reports and Examples (MTSamples) \cite{noauthor_mtsamples_nodate}, 50 from the Medical Information Mart for Intensive Care (MIMIC-III) \cite{johnson_mimic-iii_2016}, and 50 from 2010 Informatics for Integrating Biology \& the Bedside (2010 i2b2) \cite{uzuner_2010_2011}. Tables 1 presents the composition and statistics of the four datasets used in this study for NER and RE.

\begin{table}[h!]
\centering
\begin{tabular}{|l|l|c|c|c|c|c|c|c|}
\hline
 & & \multicolumn{3}{c|}{Train (Finetune)} & \multicolumn{4}{c|}{Test} \\ \hline
 & & UTP & MTSamples & MIMIC-III & UTP & MTSamples & MIMIC-III & i2b2 \\ \hline
\multicolumn{2}{|l|}{Documents}  & 1,292 & 96 & 25 & 50 & 50 & 25 & 50 \\ \hline
Main Entity & Modifier &  &  &  &  &  &  & \\ \hline
\multicolumn{2}{|l|}{Problem}  & 85,457 & 4,059 & 2,238 & 3,446 & 2,200 & 2,379 & 1,454 \\ \hline
& Negation & 46,218 & 1,478 & 438 & 1,863 & 780 & 575 & 271 \\ \hline
& Temporal & 2,737 & 101 & 156 & 114 & 65 & 181 & 73 \\ \hline
& Severity & 1,544 & 114 & 91 & 53 & 38 & 131 & 37 \\ \hline
& Condition & 296 & 17 & 42 & 18 & 20 & 74 & 1 \\ \hline
& Certainty & 551 & 62 & 87 & 15 & 20 & 63 & 12 \\ \hline
& Subject & 2,575 & 26 & 20 & 132 & 12 & 26 & 0 \\ \hline
& Body location & 24,205 & 959 & 720 & 956 & 588 & 726 & 468 \\ \hline
& Course & 2,971 & 220 & 163 & 114 & 109 & 184 & 93 \\ \hline
\multicolumn{2}{|l|}{Test}  & 50,003 & 1,529 & 2,297 & 2,151 & 828 & 2,542 & 1,309 \\ \hline
& Negation & 224 & 11 & 9 & 5 & 7 & 11 & 2 \\ \hline
& Temporal & 24,027 & 115 & 2,130 & 1,055 & 30 & 3,057 & 474 \\ \hline
& Lab value & 33,009 & 893 & 1,606 & 1,433 & 398 & 1,963 & 919 \\ \hline
& Reference range & 2,287 & 0 & 11 & 44 & 0 & 5 & 0 \\ \hline
\multicolumn{2}{|l|}{Drug}  & 13,648 & 747 & 1,047 & 568 & 466 & 1,021 & 681 \\ \hline
& Negation & 321 & 20 & 7 & 6 & 8 & 15 & 3 \\ \hline
& Temporal & 16,109 & 46 & 44 & 728 & 14 & 69 & 60 \\ \hline
& Strength & 6,590 & 154 & 492 & 275 & 171 & 436 & 268 \\ \hline
& Dosage & 5,352 & 61 & 258 & 231 & 46 & 248 & 99 \\ \hline
& Duration & 514 & 27 & 77 & 21 & 11 & 27 & 32 \\ \hline
& Form & 12,028 & 58 & 396 & 536 & 47 & 359 & 144 \\ \hline
& Frequency & 6,440 & 209 & 521 & 262 & 219 & 434 & 320 \\ \hline
& Route & 7,079 & 106 & 377 & 301 & 92 & 353 & 260 \\ \hline
\multicolumn{2}{|l|}{Other Treatment}  & 8,208 & 584 & 542 & 329 & 388 & 510 & 405 \\ \hline
& Negation & 321 & 12 & 16 & 6 & 5 & 19 & 6 \\ \hline
& Temporal & 1,534 & 53 & 56 & 59 & 38 & 98 & 59 \\ \hline
\end{tabular}
\caption{Numbers of documents, entities, and combinations between main and modifier entities for train and test splits across all datasets.}
\end{table}

\subsection*{2.3 Models}
\subsubsection*{2.3.1 Large Language Model Meta AI (LLaMA)}
We instruction-tuned LLaMA-2-chat and LLaMA-3-instruct models (7B, 13B, and 70B for LLaMA-2-chat; 8B and 70B for LLaMA-3-instruct) using Parameter-Efficient Fine Tuning (PEFT) to enhance computational efficiency \cite{touvron_llama_2023}. Specifically, we used 4-bit quantized models with QLoRA (LoRA settings: r=16, $\alpha=$64, dropout=0.05) and focused on tuning linear layers to reduce computational load \cite{dettmers_qlora_2024}. Fine-tuning parameters included a learning rate of 2e-4, 2 epochs, batch size of 4, and warmup ratio of 0.05. Model weights are available at huggingface.co/meta-llama \cite{wolf_huggingfaces_2019}, and we used the vLLM package with a temperature of 0 for reproducible output generation \cite{kwon_efficient_2023}.\\

\subsubsection*{2.3.2 Bidirectional Encoder Representations from Transformers (BERT)}
We fine-tuned BioClinicalBERT for NER and BiomedBERT for RE tasks \cite{gu_domain-specific_2022,alsentzer_publicly_2019} for NER and RE tasks. Based on our preliminary experiments, these two distinct BERT models demonstrated superior performance for their respective tasks, motivating our decision to use them separately. For the NER model, we employed a learning rate of 5e-5 over 20 epochs with a batch size of 4. The RE model was trained using a learning rate of 1e-6 over 20 epochs with a batch size of 64.\\

\subsection*{2.4 Instruction format}
Our approach uses a unified format for NER and RE (see Figure 1(B)) as listed below:
NER: LLaMA models are instructed to identify main entities in clinical note sentences, tagging them with  \textless span class='entity\_type'\textgreater ... \textless /span\textgreater , where "entity\_type" defines the entity.
RE: The main entity in the input text is tagged similarly, and LLaMA models identify related modifier entities using the same span format. This approach simplifies entity and modifier recognition in a single step. Full instructions are in Supplementary Table S1.

\subsection*{2.5 Evaluation}
\subsubsection*{2.5.1 Performance and significance test}
We evaluate RE and NER performance using Precision (P), Recall (R), and F1 scores under exact and relaxed match criteria. Exact matches require entity types, boundaries, and relations to align perfectly, while relaxed matches require only entity type and relation matches with overlapping boundaries. Significance tests assess improvements over BERT, calculating p-values via a one-sided Wilcoxon rank-sum test with bootstrapping (sample size equal to test size, 1000 repetitions, 95\% confidence interval).

\subsubsection*{2.5.2 Cross-institution generalizability}
We assess cross-institution generalizability using the i2b2 dataset, which contains clinical notes from Partners Healthcare and Beth Israel Deaconess Medical Center. This dataset's diversity makes it ideal for testing model generalization to new institutions. Therefore, we excluded i2b2 from fine-tuning to use it solely for generalizability testing.

\subsubsection*{2.5.3 Computing costs}
We assess each model's computational costs by measuring GPU hours, memory usage, and energy consumption during fine-tuning and inference. GPU hours are calculated by multiplying the number of GPUs by the total processing time. Memory usage reflects the average GPU memory in GB per GPU. Energy consumption is estimated from average GPU power usage and converted to carbon emissions using a factor of 0.39 kg CO\textsubscript{2})/kWh \cite{noauthor_how_2020}.

\subsubsection*{2.5.4 Throughput}
We assess each model's throughput during fine-tuning and inference. For fine-tuning, we measure the total time to train and calculate seconds per note by dividing the total time by the number of notes in the dataset. For inference, we determine the time per note using the test dataset with a batch size of 100 for both LLaMA and BERT models.

\section*{3 Results}
\subsection*{3.1 Fine-tuning performance  }
Table 2 highlights the exact match performance of LLaMA-2, LLaMA-3, and BERT models, with full metrics in Supplementary Table S2. LLaMA models consistently outperformed BERT, particularly with limited fine-tuning data. With sufficient data (e.g., UTP dataset), LLaMA models show minor improvements of ~1\% on NER and 1.5-3.7\% on RE. However, with less data (e.g., MTSamples, MIMIC-III), LLaMA models show greater performance improvement compared to BERT, with 3.7-4.5\% higher F1 scores on NER and up to 12.5\% higher on RE.

\begin{table}[h!]
\centering
\begin{tabular}{llcccc}
\toprule
Task & Model & UTP & MTSamples & MIMIC-III & i2b2 (Unseen) \\
\midrule
\multirow{6}{*}{NER} 
& LLaMA-2-7B & 0.929 & 0.860 & 0.838 & 0.846 \\
& LLaMA-2-13B & \textbf{\underline{0.932}} & 0.868 & 0.847 & 0.853 \\
& LLaMA-2-70B & 0.931 & 0.871 & 0.847 & 0.860 \\
& LLaMA-3-8B & 0.929 & 0.869 & 0.843 & 0.852 \\
& LLaMA-3-70B & \textbf{\underline{0.932}} & \textbf{\underline{0.876}} & \textbf{\underline{0.855}} & \textbf{\underline{0.872}} \\
& BERT & 0.921 & 0.833 & 0.810 & 0.798 \\
\midrule
\multirow{6}{*}{RE} 
& LLaMA-2-7B & 0.916 & 0.785 & 0.823 & 0.730 \\
& LLaMA-2-13B & 0.915 & 0.793 & 0.833 & 0.725 \\
& LLaMA-2-70B & 0.918 & \textbf{\underline{0.795}} & 0.850 & 0.736 \\
& LLaMA-3-8B & 0.936 & 0.787 & \textbf{\underline{0.859}} & 0.739 \\
& LLaMA-3-70B & \textbf{\underline{0.937}} & \textbf{\underline{0.795}} & 0.858 & \textbf{\underline{0.744}} \\
& BERT & 0.898 & 0.670 & 0.808 & 0.703 \\
\bottomrule
\end{tabular}
\caption{Exact F1 scores of LLaMA-2, LLaMA-3, and BERT on NER and RE across three datasets and cross-institution generalizability test results on unseen dataset (i2b2). All results of LLaMA models are statistically significant compared to BERT based on the one-sided Wilcoxon rank-sum test.}
\end{table}

To explore the performance of LLaMA and BERT in low-resource settings, we conducted dataset-specific fine-tuning by training two models—LLaMA-3-70B and BERT—separately on the MTSamples and MIMIC-III datasets. Each model was fine-tuned exclusively on either MTSamples or MIMIC-III, then evaluated across all four test sets. Based on the results in Table 2, where LLaMA-3-70B demonstrated superior performance across datasets, we prioritized using this model for these experiments. Our findings show that LLaMA-3-70B consistently outperformed BERT across datasets in these low-resource scenarios, as shown in Supplementary Tables S3 and S4.

\subsection*{3.3 Computational resources and throughput}
We compared the computational resources and throughput of BERT, LLaMA-2, and LLaMA-3 using NVIDIA A100 80GB GPUs (Tables 3 and 4). Fine-tuning was conducted with 1 GPU for BERT and the LLaMA. For inference, BERT and the 7B, 8B, and 13B LLaMA models used 1 GPU, whereas the 70B models required 2 GPUs. Results show notable trade-offs between performance and computational efficiency across model sizes.

\subsubsection*{3.3.1 Memory usage}
Memory requirements increase substantially with model size. The 70B versions of LLaMA-2 and LLaMA-3 required 58-66 GB for training and 146 GB for inference, while BERT used only 21.3 GB and 65.7 GB, respectively. The 7B and 8B LLaMA models showed comparable memory efficiency to BERT, with training memory usage between 12.7-32.4 GB. Notably, during training, LLaMA models utilize Low-Rank Adaptation (LoRA), which reduces memory requirements compared to inference, where LoRA is not applied. This difference makes the memory needed for fine-tuning significantly lower than that required for inference. Additionally, these memory usages are heavily influenced by batch sizes; the current setup is optimized for best model performance.

\subsubsection*{3.3.2 Energy consumption and carbon emission}
Energy consumption varied widely across models, with larger models requiring substantially more resources. LLaMA-70B models were the most energy-intensive during training, using 101-125 kWh and producing 39-49 kg of CO\textsubscript{2}).  Smaller LLaMA models and BERT were more efficient, using 13-28 kWh and 5-11 kg of CO\textsubscript{2}). For inference, BERT was highly efficient at 0.03 kWh, while LLaMA models ranged from 0.14 kWh (LLaMA-3-8B) to 0.86 kWh (LLaMA-2-70B).

\begin{table}[h!]
\centering
\begin{tabular}{lcccccc}
\toprule
& \makecell{\textbf{LLaMA2-}\\ \textbf{7B}} & \makecell{\textbf{LLaMA2-}\\ \textbf{13B}} & \makecell{\textbf{LLaMA2-}\\ \textbf{70B}} & \makecell{\textbf{LLaMA3-}\\ \textbf{8B}} & \makecell{\textbf{LLaMA3-}\\ \textbf{70B}} & \makecell{\textbf{Biomed}\\ \textbf{BERT}} \\
\midrule
\textbf{\textit{Memory Usage (GB)}} & & & & & & \\
Train & 12.7 & 17.4 & 57.6 & 32.4 & 66.3 & 21.3 \\
Test* & 72.8 & 72.9 & 146.9 & 72.7 & 145.9 & 65.7 \\
\midrule
\textbf{\textit{Total GPU Hours (hours)}} & & & & & & \\
Train & 41.7 & 71.2 & 320.2 & 38.6 & 273.5 & 96.8 \\
Test* & 0.5 & 0.9 & 2.2 & 0.4 & 2.2 & 0.08 \\
\midrule
\textbf{\textit{GPU Hours per Epoch (hours)}} & & & & & & \\
Train & 20.85 & 35.6 & 160.1 & 19.3 & 136.75 & 4.84 \\
\midrule
\textbf{\textit{Energy Consumption (kWh)}} & & & & & & \\
Train & 15.22 & 27.98 & 125.21 & 13.16 & 101.21 & 15.7 \\
Test* & 0.19 & 0.36 & 0.86 & 0.14 & 0.81 & 0.03 \\
\midrule
\textbf{\textit{Carbon Emissions (kg CO\textsubscript{2})}} & & & & & & \\
Train & 5.94 & 10.91 & 48.83 & 5.13 & 39.47 & 6.12 \\
Test* & 0.08 & 0.14 & 0.33 & 0.06 & 0.32 & 0.01 \\
\bottomrule
\end{tabular}
\begin{tablenotes}[flushleft]  
    \footnotesize
    \item Training: 1413 documents; Test: 175 documents. *70B models need two A100 80GB GPUs for testing.
\end{tablenotes}
\caption{Training and test times, and processing throughputs across models and datasets.}
\label{table:resource_usage}
\end{table}

\subsubsection*{3.3.3 Training and Inference Throughput}
Training and inference throughputs varied substantially across models and datasets as shown in table 4. LLaMA-3-8B had the fastest training time at 38.6 GPU hours, while BERT required 96.8 hours. BERT’s longer training time can be attributed to its need for 20 training epochs, compared to only 2 epochs for LLaMA models. This difference is reflected in GPU hours per epoch: BERT stands out with a particularly low requirement of 4.84 GPU hours per epoch, whereas LLaMA2-7B and LLaMA3-8B require 20.85 and 19.3, respectively. However, LLaMA-70B models took much longer, requiring 273.5-320.2 hours. Dataset characteristics such as note length impacted the training time; for example, MIMIC-III (2,106 tokens) took significantly more time per note than MTSamples (603 tokens). BERT outperformed all LLaMA models in inference, processing notes 5-28 times faster. Among LLaMA models, LLaMA-3-8B was quickest, while the 70B models were slowest. These findings highlight a trade-off: larger LLaMA models offer higher accuracy but are slower, while BERT is more efficient for time-sensitive applications.

\begin{table}[h!]
\centering
\begin{tabular}{lcccccc}
\toprule
& \makecell{\textbf{LLaMA2-}\\ \textbf{7B}} & \makecell{\textbf{LLaMA2-}\\ \textbf{13B}} & \makecell{\textbf{LLaMA2-}\\ \textbf{70B}} & \makecell{\textbf{LLaMA3-}\\ \textbf{8B}} & \makecell{\textbf{LLaMA3-}\\ \textbf{70B}} & \makecell{\textbf{Biomed}\\ \textbf{BERT}} \\
\midrule
\textbf{\textit{Total Training time (hours)}} & & & & & & \\
UTP & 38.3 & 65.5 & 294.5 & 35.5 & 251.5 & 89.0 \\
MTSamples & 1.8 & 3.0 & 13.5 & 1.6 & 11.5 & 4.1 \\
MIMIC-III & 1.6 & 2.7 & 12.3 & 1.5 & 10.5 & 3.7 \\
\midrule
\textbf{\textit{Training GPU hours per epoch (hours)}} & & & & & & \\
UTP & 19.2 & 32.8 & 147.3 & 17.8 & 125.8 & 4.5 \\
MTSamples & 0.9 & 1.5 & 6.8 & 0.8 & 5.8 & 0.2 \\
MIMIC-III & 0.8 & 1.4 & 6.2 & 0.8 & 5.3 & 0.2 \\
\midrule
\textbf{\textit{Training time per note (seconds/note)}} & & & & & & \\
UTP & 106.9 & 182.4 & 820.5 & 98.9 & 700.8 & 248.0 \\
MTSamples & 65.8 & 112.3 & 505.2 & 60.9 & 431.5 & 152.7 \\
MIMIC-III & 229.9 & 392.6 & 1,765.4 & 212.8 & 1,507.9 & 533.7 \\
\midrule
\textbf{\textit{Total Inference time (seconds)}}  & & & & & & \\
UTP & 725.7 & 1,271.4 & 3,009.2 & 582.1 & 2,980.9 & 145.8 \\ 
MTSamples & 266.0 & 468.5 & 1,157.9 & 232.4 & 1,142.0 & 22.3 \\
MIMIC-III & 471.0 & 815.8 & 1,998.6 & 406.9 & 2,013.9 & 47.2 \\
i2b2 & 398.8 & 695.4 & 1,753.5 & 335.4 & 1,786.2 & 81.6 \\
\midrule
\textbf{\textit{Inference timer per note (seconds/note)}} & & & & & & \\
UTP & 14.5 & 25.4 & 60.2 & 11.6 & 59.6 & 2.9 \\
MTSamples & 5.3 & 9.4 & 23.2 & 4.6 & 22.8 & 0.4 \\
MIMIC-III & 18.8 & 32.6 & 79.9 & 16.3 & 80.6 & 1.9 \\
i2b2 & 8.0 & 13.9 & 35.1 & 6.7 & 35.7 & 1.6 \\

\bottomrule
\end{tabular}
\begin{tablenotes}[flushleft]  
    \footnotesize
    \item Average training note lengths: UTP: 978 tokens/note; MTSamples: 603 tokens/note; MIMIC-III: 2106  tokens/note. Average test note lengths: UTP: 1,022 tokens/note; MTSamples: 655 tokens/note; MIMIC-III: 2,086 tokens/note; i2b2: 822 tokens/note. *70B models need two A100 80GB GPUs for inference.
\end{tablenotes}
\caption{Training and test times, and processing throughputs across models and datasets.}
\end{table}

\subsection*{3.4 Error analysis}
Automatic and manual error analyses on LLaMA and BERT outputs across four datasets revealed that LLaMA-70B models have fewer False Negatives (FN) but slightly higher False Positives (FP) than smaller models, indicating a precision-recall trade-off with model size. BERT, while showing higher FN and FP rates, has fewer boundary errors, suggesting better accuracy in detecting text boundaries. Type errors were consistently low for all models, confirming effective classification once detected. Manual analysis of 40 sentences per dataset found a 7.5\% annotation error rate, where models identified entities missed by human annotators, indicating their potential to improve pre-annotation processes.

\subsection*{3.5 Package Implementation}
To support real-world application, we created an open-source clinical IE tool, "Kiwi," featuring our LLaMA-based and BERT-based models. Kiwi, available at \url{https://kiwi.clinicalnlp.org/}, is Python-based and packaged as a Docker image for cross-platform compatibility. It includes a CPU version with BERT and a GPU version with both BERT and LLaMA models. Kiwi docker image supports batch processing, and an online demo with a GUI is also provided on the website for testing purposes (see Figure 2 for interface screenshots).

\begin{figure}[H]
  \centering
  \fbox{\includegraphics[width=1.0\textwidth]{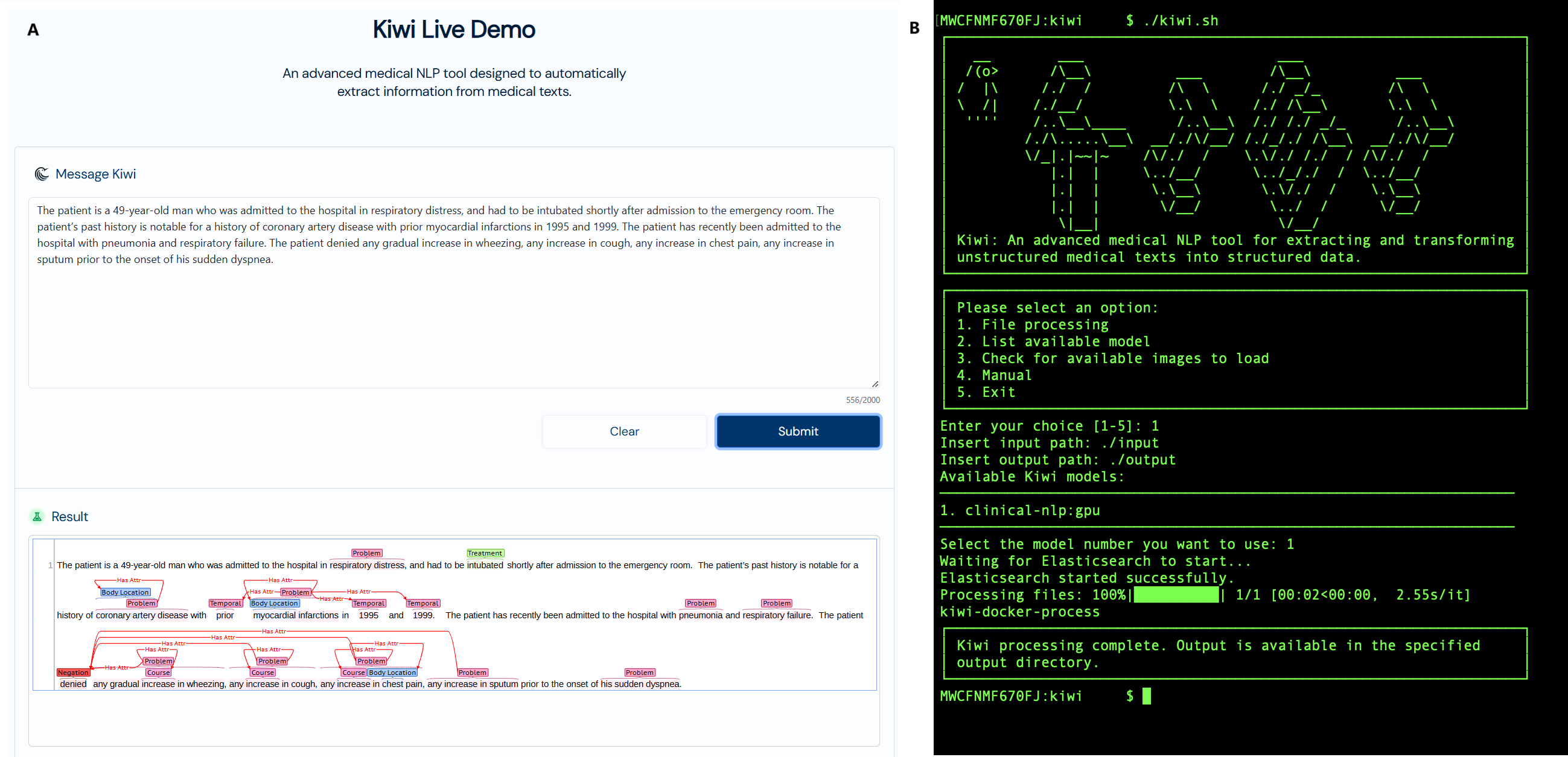}}
  \caption{Screenshots of Kiwi's Web Interface (A) and Command Line Interface (B).}
  \label{interface.png}
\end{figure}

\subsection*{4 Discussion}
This study presents the first comprehensive evaluation of instruction-tuned LLMs for clinical IE tasks using diverse institutional datasets, comparing them with previous SOTA BERT model. We address several limitations in existing LLM-based clinical IE research: previous studies often focused on datasets with limited entity types and relation types \cite{andrew_evaluating_2024,fornasiere_medical_2024}, used small corpora for evaluation \cite{agrawal_large_2022}, and primarily tested on similar types of clinical texts \cite{goel_llms_2023,andrew_evaluating_2024}. By developing a broad, multi-institutional corpus including a wide range of medical contexts, entity types, and relation types, we offer a more comprehensive and generalized assessment of LLMs' capabilities in real-world clinical IE scenarios. \\
\\
Generative AI, especially LLMs, have gained immense popularity and have opened up vast potential across various industries. In healthcare, there is growing interest in utilizing LLMs to develop and implement IE systems that support both clinical research and practice. However, our study highlights that several challenges still impede the broad adoption of LLMs for clinical IE tasks. Despite their impressive performance, issues such as high computational costs and slow processing throughputs remain significant barriers. In the following sections, we discuss how to choose between LLMs and traditional deep learning approaches for clinical IE applications, factoring in performance, throughput requirements, and the availability of computational resources. \\
\\
\textbf{Performance and Generalizability:} The consistent outperformance of LLaMA models, especially on the low-resource datasets (MTSamples, and MIMIC-III) and on unseen dataset (i2b2), underscores their potential to address the critical challenge of variability in clinical documentation \cite{kannampallil_considering_2011}. This variability, stemming from factors such as local practices, specializations, and patient demographics, has hindered the widespread adoption of NLP solutions in healthcare. The ability of instruction-tuned LLaMA models to generalize effectively across these variations suggests a promising path towards more universally applicable clinical NLP tools.\\
\\
However, it's important to note that while the performance gains are significant, they are not uniform across all tasks and datasets. When abundant annotated data are available at a healthcare system (e.g., UTP), the performance improvements of LLaMA are modest when compared with the BERT model (0.932 vs. 0.921 in F1 for NER and 0.937 vs 0.898 in F1 for RE), suggesting that deep learning models like BERT remain competitive in this scenario. Considering the low cost and high throughput of BERT models, we would recommend BERT models over LLMs when substantial annotated training data is already available for the IE task. However, when annotated training data is limited to a healthcare system (e.g., the case of i2b2), LLMs could be considered first, as they show better generalizability, indicating less annotation is needed to build the IE system. This highlights the need for careful consideration of the specific use case and available data when selecting a model for deployment.\\
\\
\textbf{Throughput:} Our study highlights a significant difference in processing throughput between LLM-based and BERT-based systems, which is crucial for real-world clinical IE applications. In settings where real-time processing is required, such as during patient care, slow NLP systems are impractical as users cannot wait for delayed results. Similarly, in research environments handling millions of clinical notes, the lower processing throughput of LLMs makes them less feasible, even if they offer slightly better performance. Ultimately, the choice between LLMs and BERT-based systems must consider both performance and throughput, as the latter is often task-specific and essential in many clinical and research scenarios.\\
\\
\textbf{Computing Resources}: The comparative analysis of computational requirements reveals important considerations for both inference and training. For inference deployment scenarios: (1) In high-resource settings with multiple NVIDIA A100 GPUs or better, LLaMA-3-70B offers the best performance; (2) In balanced settings with a single A100 GPU or equivalent, LLaMA-3-8B provides an attractive middle ground, achieving significant performance improvements; (3) For resource-constrained settings, BERT remains viable, requiring less resources while maintaining competitive performance on in-domain datasets. For training considerations, institutions have several options: (1) Using parameter-efficient tuning on larger LLaMA models (70B) demands significant resources but achieves the best performance; (2) Training smaller LLaMA models (7B/8B/13B) offers a balanced approach with competitive results; (3) Traditional fine-tuning of BERT provides the most resource-efficient though with lower performance on cross-institutional data. For institutions with extensive computational resources, pre-training domain-specific models like Me-LLaMA represents another option, though this requires more, making it practical only for large-scale research environments \cite{xie_me_2024}. These trade-offs underscore the importance of carefully matching computational capabilities with specific use case requirements.\\
\\
Based on these findings, we made some recommendations for model choice for different clinical IE tasks in different resource settings, summarized in Figure 3.

\begin{figure}[H]
  \centering
  \fbox{\includegraphics[width=1.0\textwidth]{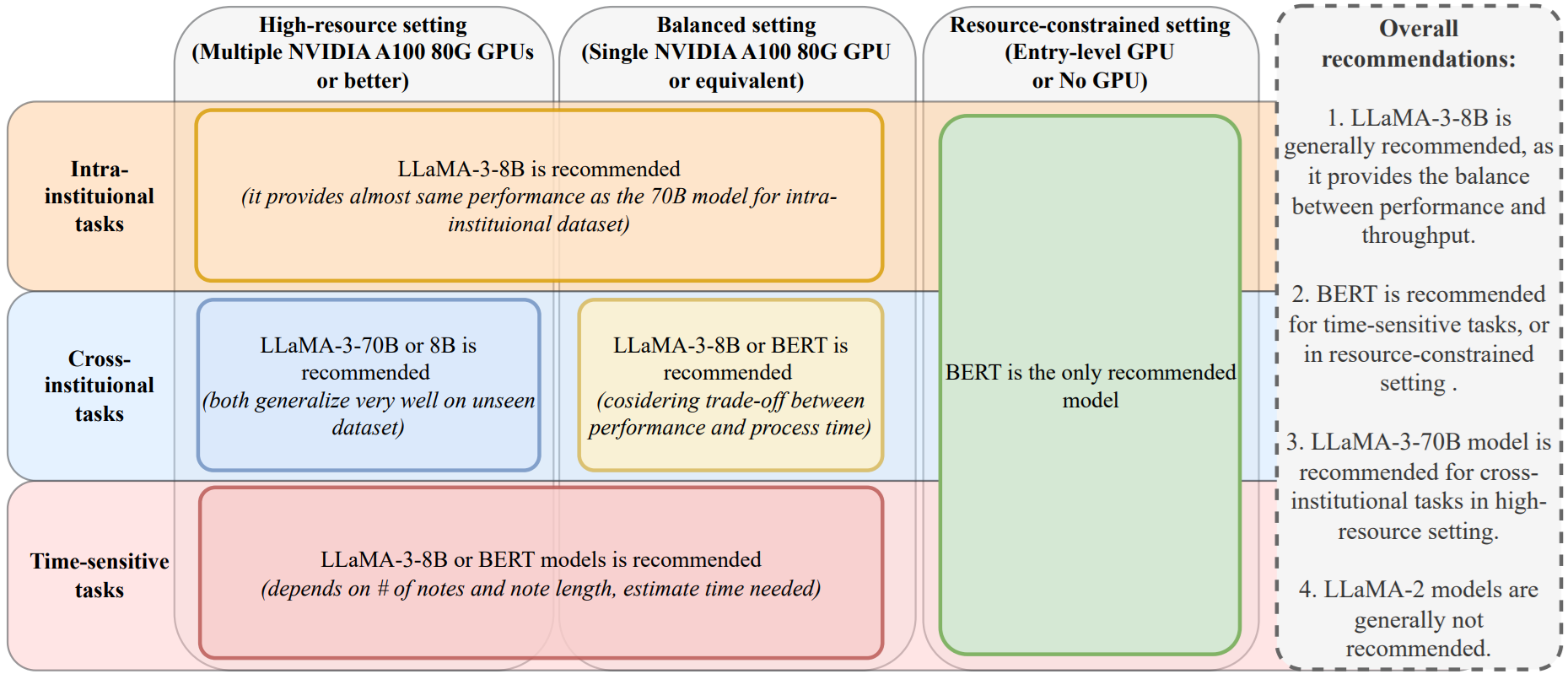}}
  \caption{Recommendations for model choice for different tasks in different resource settings.}
  \label{recommendation.png}
\end{figure}

Another significant contribution of this study is to release the Kiwi package. To the best of our knowledge, Kiwi is the first open-source clinical IE system based on LLMs. Kiwi includes both our instruction-tuned LLaMA-based and BERT-based models, available in both CPU and GPU versions to accommodate the diverse computational resources of different healthcare institutions. By offering different versions of models, Kiwi enables institutions to select the most appropriate model based on their specific needs, available resources, and use cases according to our suggestions. As a pioneering LLM-based clinical IE system, Kiwi represents a significant milestone in the application of generative AI to clinical IE. By making our system publicly available, we aim to accelerate research in LLM-based clinical IE by providing researchers with a ready-to-use framework. This will enable the medical NLP community to use our models directly, customize models upon our work, and develop more advanced clinical IE solutions. \\
\\
While our study provides valuable insights into the potential of LLMs for clinical IE, future research should explore techniques to distill the knowledge of large LLMs into more compact, resource-efficient models \cite{latif_knowledge_2024,xu_survey_2024}. This could help bridge the gap between the performance of large LLMs and the resource constraints of many healthcare settings. Future work should also explore how continual training on institution-specific datasets could further improve performance. This approach could allow individual healthcare institutions to refine the models' performance on their unique documentation styles and terminologies, potentially achieving even better results than those observed in our cross-institutional evaluation. Several limitations of our study should be noted. First, our current information extraction approach operates at the sentence level, requiring clinical notes to be segmented into individual sentences before processing. This approach may not fully leverage one of the key advantages of LLMs: their ability to handle longer sequences. Future research could explore IE at the section or document level, potentially capturing more complex cases such as inter-sentence relationships. Second, our study utilized LLaMA-2/3 as base models, which were not specifically pretrained on medical texts. Future work could investigate instruction-tuning models based on domain-specific foundational LLMs, such as Me-LLaMA \cite{xie_me_2024}, which may yield better performance due to their specialized medical knowledge.\\
\\

\subsection*{Data availability}
The MTSamples dataset can be directly accessed on our GitHub repository. The MIMIC-III and 2010 i2b2 datasets, we are providing entity offsets and texts without the original text files on our GitHub repository.

\subsection*{Code availability}
Our annotation guidelines and code used in this study are available at: \url{https://github.com/BIDS-Xu-Lab/Kiwi-LLaMA}

\subsection*{Acknowledgments}
This study is partially supported by the following grants: NIA RF1AG072799, R01AG080429, and R01AG078154.

\subsection*{Contributions}
Y.H. and H.X. contributed to the conception and design of the study. Y.H. and X.Z. performed the experiments and analyzed the data. Y.Z. led the data annotation. Y.H., Q.C., X.Z., and H.X. contributed to the interpretation of the results. Y.H., V.J., and J.H. contributed to the development of the Kiwi package. Y.H., J.H., V.J., V.K., and R.L.W. contributed to the interface development. Y.H., Q.C., X.Z, X.J, K.R., V.K., and H.X. drafted the manuscript and revised it. All authors approved the final version of the manuscript and agreed to be accountable for all aspects of the work.

\makeatletter
\renewcommand{\@biblabel}[1]{\hfill #1.}
\makeatother

\bibliographystyle{vancouver}
\bibliography{amia}  

\newpage

\subsection*{Supplementary Information:}
\subsection*{Annotation process:}
For our annotation process, we engaged five annotators to ensure reliable data labeling for entity and relation recognition tasks. Annotator A, a medical student with extensive annotation experience, supervised the entire annotation process and helped resolve inconsistencies, making them the benchmark for annotation quality. \\
During the training phase, average F1 scores for entity and relation recognition across datasets showed the following results: \\
Annotator A vs. B: 82.5\% \\
Annotator A vs. C: 79.3\% \\
Annotator A vs. D: 78.1\% \\
Annotator A vs. E: 75.4\% \\
For the large-scale annotation tasks, each annotator will be assigned approximately 25 files per round, with 1-3 files designated for review and error analysis by Annotator A to maintain quality control. Our annotation guidelines, are available in our GitHub repository https://github.com/BIDS-Xu-Lab/Kiwi-LLaMA.\\
\\

\renewcommand{\tablename}{Supplementary Table}
\setcounter{table}{0} 
\begin{xltabular}{\textwidth}{p{0.15\textwidth}|X}
\toprule
Task	            & Prompt \\
\hline
\multirow{3}{8em}{NER} & \#\#\# Task\\
 & Your task is to generate an HTML version of an input text, using HTML  \textless span \textgreater tags to mark up specific entities. \\
 & \\ 
 & \#\#\# Entity Markup Guides: \\
 & Use  \textless span class="problem"\textgreater to denote a medical problem. \\
 & Use  \textless span class="treatment"\textgreater to denote a treatment. \\
 & Use  \textless span class="test"\textgreater to denote a test. \\
 & Use  \textless span class="drug"\textgreater to denote a drug. \\
 & \\ 
 & \#\#\# Entity Definitions: \\
 & Medical Problem: The abnormal condition that happens physically or mentally to a patient. \\
 & Treatment: The procedures, interventions, and substances given to a patient for treating a problem. \\
 & Drug: Generic or brand name of a single medication or a collective name of a group of medication. \\
 & Test: A medical procedure performed (i) to detect or diagnose a problem, (ii) to monitor diseases, disease processes, and susceptibility, or (iii) to determine a course of treatment. \\
  & \\ 
 & \#\#\# Input Text: Ortho Micronor 0.35 MG ... \\
 & \#\#\# Output Text:  \textless span class="drug"\textgreater Ortho Micronor  \textless /span\textgreater 0.35 MG ... \\

\hline
\multirow{3}{8em}{\makecell{RE for problem \\ entities}} & \#\#\# Task\\
 & Your task is to mark up modifier entities related to the entity marked with  \textless span \textgreater tag in the input text. \\
 & \\ 
 & \#\#\# Entity Markup Guides: \\
 & Use  \textless span class="uncertain"\textgreater  to denote a measure of doubt. \\
 & Use  \textless span class="condition"\textgreater  to denote a phrase that indicates the problems existing in a certain situation. \\
 & Use  \textless span class="subject"\textgreater  to denote the person entity who is experiencing the disorder. \\
 & Use  \textless span class="negation"\textgreater  to denote the phrase that indicates the absence of an entity. \\
 & Use  \textless span class="bodyloc"\textgreater  to denote the location on the body where the observation is present. \\
 & Use  \textless span class="severity"\textgreater  to denote the degree of intensity of a clinical condition. \\
 & Use  \textless span class="temporal"\textgreater  to denote a calendar date, time, or duration related to a problem. \\
 & Use  \textless span class="course"\textgreater  to denote the development or alteration of a problem. \\
   & \\ 
 & \#\#\# Input Text: probable  \textless span class="problem"\textgreater left paravertebral dilated vascular structure \textless /span\textgreater  … \\
 & \#\#\# Output Text:  \textless span class="uncertain"\textgreater probable \textless /span\textgreater  left  \textless span class="bodyloc"\textgreater paravertebral \textless /span\textgreater  dilated vascular structure … \\
 \hline
\multirow{3}{8em}{\makecell{RE for treatment \\ entities}} & \#\#\# Task\\
 & Your task is to mark up modifier entities related to the entity marked with  \textless span \textgreater tag in the input text. \\
 & \\ 
 & \#\#\# Entity Markup Guides: \\
 & Use  \textless span class="temporal"\textgreater  to denote a calendar date, time, or duration related to a treatment. \\
 & Use  \textless span class="negation"\textgreater  to denote the phrase that indicates the absence of an entity. \\
   & \\ 
 & \#\#\# Input Text: No  \textless span class="treatment"\textgreater further intervention \textless /span\textgreater  was done . \\
 & \#\#\# Output Text:  \textless span class=" negation"\textgreater No \textless /span\textgreater  further intervention was done . \\

 \hline
\multirow{3}{8em}{\makecell{RE for test \\ entities}} & \#\#\# Task\\
 & Your task is to mark up modifier entities related to the entity marked with  \textless span \textgreater tag in the input text. \\
 & \\ 
 & \#\#\# Entity Markup Guides: \\
 & Use  \textless span class="labvalue"\textgreater  to denote a numeric value or a normal description of the result of a lab test. \\
 & Use  \textless span class="reference\_range"\textgreater  to denote the range or interval of values that are deemed as normal for a test in a healthy person. \\
 & Use  \textless span class="negation"\textgreater  to denote the phrase that indicates the absence of an entity. \\
 & Use  \textless span class="temporal"\textgreater  to denote a calendar date, time, or duration related to a test. \\
   & \\ 
 & \#\#\# Input Text:  \textless span class="test"\textgreater Hgb \textless /span\textgreater  10.6 gm / dL \\
 & \#\#\# Output Text: Hgb  \textless span class="labvalue"\textgreater 10.6 gm / dL \textless /span\textgreater  \\

 \hline
\multirow{3}{8em}{\makecell{RE for drug \\ entities}} & \#\#\# Task\\
 & Your task is to mark up modifier entities related to the entity marked with  \textless span \textgreater tag in the input text. \\
 & \\ 
 & \#\#\# Entity Markup Guides: \\
 & Use  \textless span class="form"\textgreater  to denote the form of drug. \\
 & Use  \textless span class="frequency"\textgreater  to denote the frequency of taking a drug. \\
 & Use  \textless span class="dosage"\textgreater  to denote the amount of active ingredient from the number of drugs prescribed. \\
 & Use  \textless span class="duration"\textgreater  to denote the time period a patient should take a drug. \\
 & Use  \textless span class="strength"\textgreater  to denote the amount of active ingredient in a given dosage form. \\
 & Use  \textless span class="route"\textgreater  to denote the way by which a drug, fluid, poison, or other substance is taken into the body. \\
 & Use  \textless span class="negation"\textgreater  to denote the phrase that indicates the absence of an entity. \\
 & Use  \textless span class="temporal"\textgreater  to denote a calendar date, time, or duration related to a drug. \\
   & \\ 
 & \#\#\# Input Text: \textless span class="drug"\textgreater Ortho Micronor \textless /span\textgreater  0.35 MG ... \\
 & \#\#\# Output Text: Ortho Micronor  \textless span class="strength"\textgreater 0.35 MG \textless /span\textgreater  ... \\
\bottomrule
\caption{Complete instruction examples for NER and RE}
\end{xltabular}

\renewcommand{\tablename}{Supplementary Table}
\setcounter{table}{1} 
\setlength{\tabcolsep}{4pt} 
\begin{xltabular}{\textwidth}{|l|l|*{12}{c|}}
\hline
\textbf{Dataset} & \textbf{Model} & \multicolumn{6}{c|}{\textbf{NER}} & \multicolumn{6}{c|}{\textbf{RE}} \\
\cline{3-14}
 & & \multicolumn{3}{c|}{Exact Match} & \multicolumn{3}{c|}{Relax Match} & \multicolumn{3}{c|}{Exact Match} & \multicolumn{3}{c|}{Relax Match} \\
\cline{3-14}
 & & P & R & F1 & P & R & F1 & P & R & F1 & P & R & F1 \\
\hline
UTP & LLaMA-2-7b & 0.929 & 0.930 & 0.929 & 0.962 & 0.964 & 0.963 & 0.911 & 0.922 & 0.916 & 0.926 & 0.937 & 0.931 \\
 & LLaMA-2-13b & 0.930 & 0.934 & \textbf{\underline{0.932}} & 0.961 & 0.967 & 0.964 & 0.903 & 0.928 & 0.915 & 0.917 & 0.944 & 0.930 \\
 & LLaMA-2-70b & 0.927 & 0.934 & 0.931 & 0.960 & 0.967 & 0.964 & 0.902 & 0.933 & 0.918 & 0.917 & 0.949 & 0.933 \\
 & LLaMA-3-8b & 0.933 & 0.926 & 0.929 & 0.968 & 0.961 & \textbf{\underline{0.965}} & 0.930 & 0.943 & 0.936 & 0.944 & 0.959 & 0.952 \\
 & LLaMA-3-70b & 0.928 & 0.936 & \textbf{\underline{0.932}} & 0.960 & 0.969 & 0.964 & 0.925 & 0.949 & \textbf{\underline{0.937}} & 0.940 & 0.965 & \textbf{\underline{0.953}} \\
 & BERT & 0.918 & 0.924 & 0.921 & 0.953 & 0.962 & 0.957 & 0.853 & 0.949 & 0.898 & 0.866 & 0.967 & 0.914 \\
\hline
MTSamples & LLaMA-2-7b & 0.854 & 0.867 & 0.860 & 0.916 & 0.931 & 0.923 & 0.787 & 0.783 & 0.785 & 0.865 & 0.862 & 0.864 \\
 & LLaMA-2-13b & 0.861 & 0.874 & 0.868 & 0.921 & 0.936 & 0.928 & 0.773 & 0.814 & 0.793 & 0.849 & 0.897 & 0.872 \\
 & LLaMA-2-70b & 0.863 & 0.879 & 0.871 & 0.919 & 0.938 & 0.928 & 0.770 & 0.823 & \textbf{\underline{0.795}} & 0.848 & 0.907 & 0.877 \\
 & LLaMA-3-8b & 0.869 & 0.869 & 0.869 & 0.931 & 0.931 & 0.931 & 0.774 & 0.801 & 0.787 & 0.854 & 0.883 & 0.868 \\
 & LLaMA-3-70b & 0.868 & 0.885 & \textbf{\underline{0.876}} & 0.924 & 0.944 & \textbf{\underline{0.934}} & 0.771 & 0.820 & \textbf{\underline{0.795}} & 0.857 & 0.910 & \textbf{\underline{0.883}} \\
 & BERT & 0.830 & 0.836 & 0.833 & 0.904 & 0.915 & 0.910 & 0.570 & 0.815 & 0.670 & 0.634 & 0.920 & 0.751 \\
\hline
MIMIC-III & LLaMA-2-7b & 0.841 & 0.835 & 0.838 & 0.931 & 0.921 & 0.926 & 0.892 & 0.764 & 0.823 & 0.951 & 0.815 & 0.878 \\
 & LLaMA-2-13b & 0.846 & 0.849 & 0.847 & 0.932 & 0.934 & 0.933 & 0.880 & 0.790 & 0.833 & 0.942 & 0.843 & 0.890 \\
 & LLaMA-2-70b & 0.843 & 0.850 & 0.847 & 0.929 & 0.936 & 0.933 & 0.884 & 0.819 & 0.850 & 0.943 & 0.872 & 0.906 \\
 & LLaMA-3-8b & 0.849 & 0.838 & 0.843 & 0.936 & 0.923 & 0.930 & 0.898 & 0.824 & \textbf{\underline{0.859}} & 0.948 & 0.869 & \textbf{\underline{0.907}} \\
 & LLaMA-3-70b & 0.854 & 0.856 & \textbf{\underline{0.855}} & 0.938 & 0.940 & \textbf{\underline{0.939}} & 0.901 & 0.820 & 0.858 & 0.955 & 0.866 & 0.908 \\
 & BERT & 0.814 & 0.805 & 0.810 & 0.919 & 0.903 & 0.911 & 0.777 & 0.842 & 0.808 & 0.823 & 0.900 & 0.860 \\
\hline
i2b2 & LLaMA-2-7b & 0.844 & 0.848 & 0.846 & 0.922 & 0.920 & 0.921 & 0.783 & 0.683 & 0.730 & 0.873 & 0.749 & 0.806 \\
 & LLaMA-2-13b & 0.848 & 0.859 & 0.853 & 0.923 & 0.927 & 0.925 & 0.757 & 0.694 & 0.725 & 0.848 & 0.766 & 0.805 \\
 & LLaMA-2-70b & 0.854 & 0.867 & 0.860 & 0.922 & 0.930 & 0.926 & 0.749 & 0.723 & 0.736 & 0.839 & 0.798 & 0.818 \\
 & LLaMA-3-8b & 0.857 & 0.847 & 0.852 & 0.936 & 0.917 & 0.926 & 0.761 & 0.719 & 0.739 & 0.856 & 0.796 & 0.825 \\
 & LLaMA-3-70b & 0.870 & 0.874 & \textbf{\underline{0.872}} & 0.932 & 0.931 & \textbf{\underline{0.932}} & 0.761 & 0.727 & \textbf{\underline{0.744}} & 0.853 & 0.804 & \textbf{\underline{0.828}} \\
 & BERT & 0.791 & 0.805 & 0.798 & 0.887 & 0.905 & 0.896 & 0.649 & 0.766 & 0.703 & 0.730 & 0.849 & 0.785 \\
\hline
\caption{Exact and relax performance by LLaMA-2, LLaMA-3, and BERT on NER and RE across four datasets.}
\end{xltabular}

\renewcommand{\tablename}{Supplementary Table}
\setcounter{table}{2} 
\setlength{\tabcolsep}{4pt} 
\begin{xltabular}{\textwidth}{|l|l|*{12}{c|}}
\hline
\textbf{Dataset} & \textbf{Model} & \multicolumn{6}{c|}{\textbf{NER}} & \multicolumn{6}{c|}{\textbf{RE}} \\
\cline{3-14}
 & & \multicolumn{3}{c|}{Exact Match} & \multicolumn{3}{c|}{Relax Match} & \multicolumn{3}{c|}{Exact Match} & \multicolumn{3}{c|}{Relax Match} \\
\cline{3-14}
 & & P & R & F1 & P & R & F1 & P & R & F1 & P & R & F1 \\
\hline
UTP & LLaMA-2-7b & 0.781 & 0.743 & \textbf{\underline{0.761}} & 0.875 & 0.852 & \textbf{\underline{0.863}} & 0.820 & 0.650 & \textbf{\underline{0.725}} & 0.890 & 0.704 & \textbf{\underline{0.786}} \\
 & BERT & 0.656 & 0.666 & 0.661 & 0.786 & 0.828 & 0.806 & 0.538 & 0.704 & 0.610 & 0.571 & 0.773 & 0.657 \\
\hline
MTSamples & LLaMA-2-7b & 0.792 & 0.825 & \textbf{\underline{0.808}} & 0.879 & 0.921 & \textbf{\underline{0.900}} & 0.751 & 0.711 & \textbf{\underline{0.730}} & 0.846 & 0.798 & \textbf{\underline{0.821}} \\
 & BERT & 0.796 & 0.812 & 0.804 & 0.872 & 0.906 & 0.889 & 0.499 & 0.741 & 0.596 & 0.566 & 0.853 & 0.680 \\
\hline
MIMIC-III & LLaMA-2-7b & 0.816 & 0.821 & \textbf{\underline{0.819}} & 0.915 & 0.920 & \textbf{\underline{0.918}} & 0.880 & 0.780 & \textbf{\underline{0.827}} & 0.937 & 0.832 & \textbf{\underline{0.882}} \\
 & BERT & 0.819 & 0.817 & 0.818 & 0.912 & 0.923 & 0.917 & 0.724 & 0.786 & 0.754 & 0.775 & 0.848 & 0.810 \\
\hline
i2b2 & LLaMA-2-7b & 0.786 & 0.799 & \textbf{\underline{0.793}} & 0.888 & 0.901 & \textbf{\underline{0.894}} & 0.744 & 0.643 & \textbf{\underline{0.690}} & 0.858 & 0.721 & \textbf{\underline{0.783}} \\
 & BERT & 0.761 & 0.806 & 0.783 & 0.847 & 0.918 & 0.881 & 0.589 & 0.686 & 0.634 & 0.692 & 0.793 & 0.739 \\
\hline
\caption{Exact and relax performance by LLaMA-3-70B, and BER on NER and RE across four datasets solely using MIMIC-III dataset for finetuning.}
\end{xltabular}

\end{document}